\documentclass[runningheads]{llncs}
\usepackage[T1]{fontenc}
\usepackage{graphicx}
\usepackage{color}
\usepackage[colorlinks]{hyperref}
\usepackage{graphicx}%
\usepackage{multirow}%
\usepackage{amsmath,amssymb,amsfonts}%
\usepackage{mathrsfs}%
\usepackage[title]{appendix}%
\usepackage{xcolor}%
\usepackage{textcomp}%
\usepackage{manyfoot}%
\usepackage{booktabs}%
\usepackage{algorithm}%
\usepackage{algorithmicx}%
\usepackage{algpseudocode}%
\usepackage{listings}%
\usepackage{wrapfig}
\usepackage{enumitem}

\usepackage{tikz}
\usetikzlibrary{arrows.meta, positioning}

\newcommand{\tdata}[1]{\texttt{#1}}

\newcommand{\blue}[1]{\textcolor{blue}{#1}}
\newcommand{\magenta}[1]{\textcolor{magenta}{#1}}
\newcommand{\teal}[1]{\textcolor{teal}{#1}}

\begin{document}
\title{Cross-View Topology-Aware Graph Representation Learning}
\titlerunning{Topological Contrastive Learning for Graphs}
%
\author{Ahmet Sami Korkmaz\inst{1,}\thanks{Co-First author.} \and
Selim Coskunuzer\inst{2,\star} \and
Md Joshem Uddin\inst{3,}\thanks{Senior author}}
\authorrunning{Ahmet et al.}
%
\institute{Dept. of Mathematics, Georgia Institute of Technology, Atlanta, GA, USA \\
\email{samkorkmaz@gatech.edu} \and
Dept. of Computer Science $\&$ Engineering, Texas A$\&$M University, College Station, TX, USA\\ 
\email{selim05@tamu.edu} \and
Dept. of Mathematical Sciences, The University of Texas at Dallas, Richardson, TX, USA \\
\email{mdjoshem.uddin@utdallas.edu}}
\maketitle              
\begin{abstract}Graph classification has gained significant attention due to its applications in chemistry, social networks, and bioinformatics. While Graph Neural Networks (GNNs) effectively capture local structural patterns, they often overlook global topological features that are critical for robust representation learning. In this work, we propose GraphTCL, a dual-view contrastive learning framework that integrates structural embeddings from GNNs with topological embeddings derived from persistent homology. By aligning these complementary views through a cross-view contrastive loss, our method enhances representation quality and improves classification performance. Extensive experiments on benchmark datasets, including TU and OGB molecular graphs, demonstrate that GraphTCL consistently outperforms state-of-the-art baselines. This study highlights the importance of topology-aware contrastive learning for advancing graph representation methods.

\end{abstract}
\section{Introduction}
Graphs are the backbone of modern data science, powering breakthroughs in drug discovery, social network analysis, and infrastructure resilience~\cite{olojede2024cyber}. Despite remarkable progress, most graph neural networks (GNNs) remain constrained by the locality of message passing, which aggregates information within limited neighborhoods and struggles to capture global topological structure—the very property that defines many graph families~\cite{alon2020bottleneck,horn2022topological}. Consequently, even deep or expressive GNNs may assign similar embeddings to graphs that differ fundamentally in their connectivity or cyclic structure, leading to reduced robustness and poor transferability across graph regimes.

To bridge this gap, recent studies have explored the integration of Topological Data Analysis (TDA) into graph learning. Persistent homology and related descriptors summarize multiscale connectivity and offer global, architecture-agnostic invariants that complement the locality of message passing~\cite{coskunuzer2024topological}. However, most existing approaches inject topological features as auxiliary inputs or implicit regularizers inside the GNN, blurring the distinction between structural and topological representations and limiting interpretability.

In this work, we propose to treat these two sources of information—structural and topological—as distinct but complementary views of the same graph. The structural view is captured through a standard GNN encoder that models attribute-driven local interactions, while the topological view is represented via a lightweight MLP operating on persistent-homology–derived vectors that summarize global connectivity. A central challenge is that these two views often reside in different latent spaces: message-passing embeddings reflect feature similarity, whereas topological embeddings reflect homological structure. Without explicit alignment, the model may learn inconsistent representations.

To resolve this, we introduce a dual-view contrastive learning framework that aligns the two modalities through a cross-view contrastive loss. This alignment enforces consistency between the GNN and topological embeddings while preserving their complementary information. The fused representation is then fed into a simple linear classifier for downstream prediction. Our framework thereby offers a conceptually simple yet theoretically grounded way to enhance GNN expressivity without architectural complexity or handcrafted augmentations. Our model achieves strong generalization on several benchmark datasets including TU datasets, and OGBG dataset outperforming baseline GNNs and feature-only approaches.

Our approach is also complementary to recent distributional perspectives on graph classification, which replace global pooling with comparisons in an embedding-distribution space, and to topology-aware pooling that integrates persistence into hierarchical summarization~\cite{wang2024graph}. By aligning structural and topological views before classification, we retain the simplicity of a linear head while capturing both local and global structure. 

Our main contributions are summarized as follows:

\begin{itemize}[leftmargin=2em,label={\large\textbf{$\bullet$}}]
    \item \textbf{Dual-view contrastive framework:}  
    We introduce a novel architecture for graph classification that combines structural embeddings from a GNN with topological embeddings derived from persistent homology. These two modalities are aligned via a cross-view contrastive loss and passed through a linear classifier. 

 \item \textbf{Topology-aware representation learning:}  
    Unlike prior work that injects topological features as auxiliary signals, our approach enforces explicit alignment between structural and topological spaces. This design improves interpretability and robustness while avoiding architectural complexity or handcrafted augmentations.
    
    \item \textbf{Extensive evaluation:}  
   We validate our framework on widely used graph classification benchmarks (TU datasets and OGB molecular graphs), achieving consistent improvements over state-of-the-art GNNs, graph contrastive learning baselines, and topology-aware models. Ablation studies further isolate the contributions of topology and contrastive alignment.
\end{itemize}

\section{Related Work}

\subsection{Graph Representation and Contrastive Learning}

Classical graph classification has largely relied on kernels (e.g., random walk, shortest path, Weisfeiler--Lehman) that embed graphs into a reproducing kernel Hilbert space and train a shallow classifier~\cite{Shervashidze2011WLKernel,Vishwanathan2010GraphKernels}. Although effective on small and medium benchmarks, these methods depend on hand-crafted substructure counts and often scale poorly or generalize weakly across domains.

Graph neural networks (GNNs) such as GCN, GraphSAGE, GAT, and GIN instead learn node embeddings via message passing and aggregate them into graph-level representations~\cite{Kipf2017GCN,xu2019how}. Despite strong performance, they are limited by the expressivity of the 1-Weisfeiler--Lehman test and may fail to distinguish graphs differing in higher-order structures or global topology~\cite{Sato2020ExpressiveSurvey}, motivating more expressive architectures and readouts~\cite{Lee2019SetTransformer,Yan2022NeuralTopoFeatures}. In parallel, contrastive methods (e.g., InfoGraph, GraphCL, GRACE) learn graph representations by maximizing agreement across augmented views~\cite{Sun2020InfoGraph,You2020GraphCL}, but still operate on a single structural modality, as all views arise from graph augmentations processed by the same encoder.

\subsection{Topological Methods and Topology Aware Contrastive Learning}

Topological Data Analysis (TDA), and in particular persistent homology, provides a principled way to extract multi-scale invariants (connected components, cycles, higher-order cavities) from graphs and graph signals. Early TDA-based models compute persistence diagrams (or vectorizations) and feed them to classical classifiers or MLPs in domains such as chemistry, brain networks, and infrastructure systems~\cite{carriere2020perslay}, showing that topological signatures capture global structure beyond local subgraph counts, but treating them as fixed descriptors decoupled from learned representations.

More recent work integrates topology into end-to-end graph learning: neural surrogates of topological features~\cite{Yan2022NeuralTopoFeatures} and topology-aware architectures or pooling modules inject persistent-homology–based invariants into message passing and hierarchical readout~\cite{horn2022topological}. TDA has also been combined with contrastive objectives, TopoGCL~\cite{Chen2024TopoGCL}, the most closely related to our work.

\paragraph{Position of GraphTCL.}

GraphTCL builds on prior topology-aware and contrastive methods but adopts a different alignment mechanism. TopoGCL~\cite{Chen2024TopoGCL} is the closest work to ours: it introduces a trainable extended topological layer based on extended persistent homology and proposes a \emph{topo--topo} contrastive mode that contrasts extended persistence landscapes (and images) across augmented views, together with a standard global--global contrastive loss on GNN embeddings for unsupervised graph classification. Topology thus acts as an internal self-supervised signal inside a single encoder. Other topology-aware methods, such as differentiable persistence layers and pooling modules~\cite{carriere2020perslay,horn2022topological}, similarly inject PH-derived features into one GNN pipeline rather than modeling topology as a separate representation space.

In contrast, GraphTCL takes a dual-view perspective: each graph is represented (i) structurally, by a message-passing GNN encoder, and (ii) topologically, by a lightweight MLP operating on persistent-homology–based vectors that summarize global connectivity. We then introduce a \emph{cross-view contrastive loss} that directly aligns these structural and topological embeddings in a shared latent space, enforcing consistency between local feature-driven structure and global topological summaries. Topology is therefore not only an auxiliary regularizer but a complementary modality that actively shapes the learned representation. To the best of our knowledge, no prior work explicitly aligns \emph{separate} structural and topological graph embeddings via dual-view contrastive learning, particularly in the fully supervised setting

\section{Background}

\subsection{Graph Neural Networks}  
Graph neural networks (GNNs) provide a powerful framework for learning graph representations by propagating and aggregating information across local neighborhoods. Although our architecture can be integrated with any GNN model, in this work we adopt the Graph Isomorphism Network (GIN)~\cite{xu2019how}, one of the most expressive message-passing architectures. Given a graph $\mathcal{G}=(\mathcal{V},\mathcal{E})$ with node features $x_v=h_{v}^{(0)}$, the GIN update at layer $k$ is defined as
\[
h_v^{(k)} = \text{MLP}^{(k)}\!\Big((1+\epsilon) \cdot h_v^{(k-1)} + \sum_{u \in \mathcal{N}(v)} h_u^{(k-1)}\Big),
\]
where $\mathcal{N}(v)$ denotes neighbors of node $v$ and $\epsilon$ is a learnable parameter. After several layers of aggregation, we apply \emph{global mean pooling} to obtain a graph-level embedding. 
For datasets without node features, such as IMDB, we use node degrees as input features.

\subsection{Persistent Homology (PH)}  
Persistent Homology (PH) provides a rigorous mathematical framework for uncovering hidden topological structures in complex data, drawing on tools from algebraic topology and applied across diverse data modalities, including point clouds, images, and networks~\cite{coskunuzer2024topological}. In this section, we briefly describe PH in the context of graphs.

Let $\mathcal{G} = \{\mathcal{V}, \mathcal{E}, \mathcal{W}\}$ denote a weighted graph with node set $\mathcal{V}$, edge set $\mathcal{E}$, and weight matrix $\mathcal{W} = [\omega_{rs}]_{1 \leq r,s \leq N}$, where $\omega_{rs} > 0$ if $(v_r, v_s) \in \mathcal{E}$ and $\omega_{rs} = 0$ otherwise. For unweighted networks, we set $\omega_{rs} = 1$ for all existing edges and $0$ otherwise. The PH machinery proceeds in three main steps: \emph{filtration}, \emph{persistence diagram computation}, and \emph{vectorization}.

\smallskip
\noindent\textbf{Step 1 -- Filtration.}  
The first step is to build a \emph{filtration}, a nested sequence of simplicial complexes capturing how topology evolves. Given a node-based filtration function $f:\mathcal{V}\rightarrow\mathbb{R}$ and thresholds $\alpha_1 < \cdots < \alpha_N$, we form subgraphs
\[
\mathcal{G}^i = (\mathcal{V}_i,\mathcal{E}_i), \quad 
\mathcal{V}_i = \{v \in \mathcal{V} \mid f(v)\le \alpha_i\}, \ 
\mathcal{E}_i = \{(u,v)\in\mathcal{E}\mid u,v\in\mathcal{V}_i\},
\]
yielding a sublevel filtration $\mathcal{G}^1 \subset \cdots \subset \mathcal{G}^N=\mathcal{G}$. Each $\mathcal{G}^i$ is then lifted (e.g., via the clique complex) to a simplicial complex $\widehat{\mathcal{G}}^i$, producing a nested sequence $\widehat{\mathcal{G}}^1 \subseteq \cdots \subseteq \widehat{\mathcal{G}}^N$ that tracks the emergence of higher-order structures.


\smallskip
\noindent\textbf{Step 2 -- Persistence Diagrams.}  

Once the filtration is constructed, PH tracks the birth and death of topological features across the sequence of complexes. A $k$-dimensional feature corresponds to a connected component ($k=0$), loop ($k=1$), or void ($k=2$). For each feature $\sigma$, PH records its \emph{birth} $b_\sigma$ and \emph{death} $d_\sigma$, yielding the $k$-dimensional persistence diagram \quad
$\mathrm{PD}_k(\mathcal{G}) = \{(b_\sigma, d_\sigma) \mid b_\sigma < d_\sigma\},$ \quad 
where points far from the diagonal ($d_\sigma - b_\sigma$ large) are typically regarded as topologically significant. These diagrams serve as compact topological fingerprints of the graph.

\smallskip
\noindent\textbf{Step 3 -- Vectorization.}  
While persistence diagrams encode rich structural information, they are multisets of points in $\mathbb{R}^2$ and thus not directly compatible with standard learning pipelines. They are therefore transformed into fixed-length representations via \emph{vectorization} or \emph{kernelization}~\cite{ali2023survey}. Common choices include persistence images, landscapes, and Betti curves, which summarize the lifetimes and distributions of topological features in a compact form suitable for statistical and deep learning models.


\paragraph{Heat Kernel Signature (HKS) as a Filtration Function.}
The Heat Kernel Signature (HKS)~\cite{carriere2020perslay} provides an intrinsic multi-scale descriptor of a graph based on heat diffusion governed by the graph Laplacian $L = D - A$. The heat kernel $H_t = e^{-tL}$ has diagonal entries $H_t(v,v)$ measuring the heat retained at node $v$ after time $t$. Using the spectral decomposition $L \phi_i = \lambda_i \phi_i$, the HKS at node $v$ and time $t$ is
\[
f(v,t) = \sum_{i} e^{-\lambda_i t} \phi_i(v)^2,
\]
where $\{\lambda_i\}$ and $\{\phi_i\}$ denote the eigenvalues and corresponding eigenfunctions of $L$, respectively. In our framework, HKS values serve as the \emph{filtration function} for persistent homology, so that topological features reflect the intrinsic diffusion geometry of the graph and complement the learned structural embeddings.


\section{GraphTCL: Topological Contrastive Learning}  
The core idea of our framework is to jointly leverage structural and topological information by aligning them in a shared embedding space. Our \textbf{GraphTCL} integrates: (a) a GIN or GCN encoder for structural embeddings ($z_i$), (b) an MLP encoder for topological embeddings ($u_i$), and (c) a fusion module that concatenates $z_i$ and $u_i$, projects them into a hidden layer, and outputs both (i) class logits and (ii) a normalized projection vector for contrastive learning.

Formally, for graph $i$ we compute
\[
z_i = \phi_{\text{GNN}}(x_i, \mathcal{E}_i), \quad
u_i = \phi_{\text{MLP}}(t_i),
\]
\[
f_i = \sigma(W [z_i \Vert u_i]), \quad
y_i = \text{softmax}(W_c f_i), \quad
p_i = \text{Proj}(f_i),
\]
where $f_i$ is the fused embedding, $y_i$ is the predicted class label, and $p_i$ is the projection used in the contrastive objective.
\paragraph{Contrastive Loss.}  
Given a minibatch of $B$ graphs, we obtain structural embeddings 
$\{z_i\}_{i=1}^B$ from the GNN encoder and topological embeddings 
$\{u_i\}_{i=1}^B$ from the MLP encoder. 
Each pair $(z_i, u_i)$ forms a \emph{positive pair}, while 
$(z_i, u_j), \, j \neq i$ and $(u_i, z_j), \, j \neq i$ are treated 
as \emph{negatives}. The bidirectional InfoNCE objective is then
\[
\mathcal{L}_{\text{con}} = 
-\frac{1}{2B} \sum_{i=1}^B 
\Bigg[
\log \frac{\exp(S(z_i,u_i))}{\sum_{j=1}^B \exp(S(z_i,u_j))} \;+\;
\log \frac{\exp(S(u_i,z_i))}{\sum_{j=1}^B \exp(S(u_i,z_j))}
\Bigg].
\]
where $S(v_i,v_j)=\frac{\langle v_i, v_j \rangle}{\tau}$ define the cosine similarity with temperature parameter $\tau > 0$.  This formulation enforces that structural and topological embeddings 
of the same graph are close in the shared space, while embeddings of 
different graphs remain apart. Importantly, the two terms correspond 
to the two retrieval directions: (\emph{structure $\rightarrow$ topology}) 
and (\emph{topology $\rightarrow$ structure}).



\begin{figure}
    \centering
    \includegraphics[width=.8\linewidth]{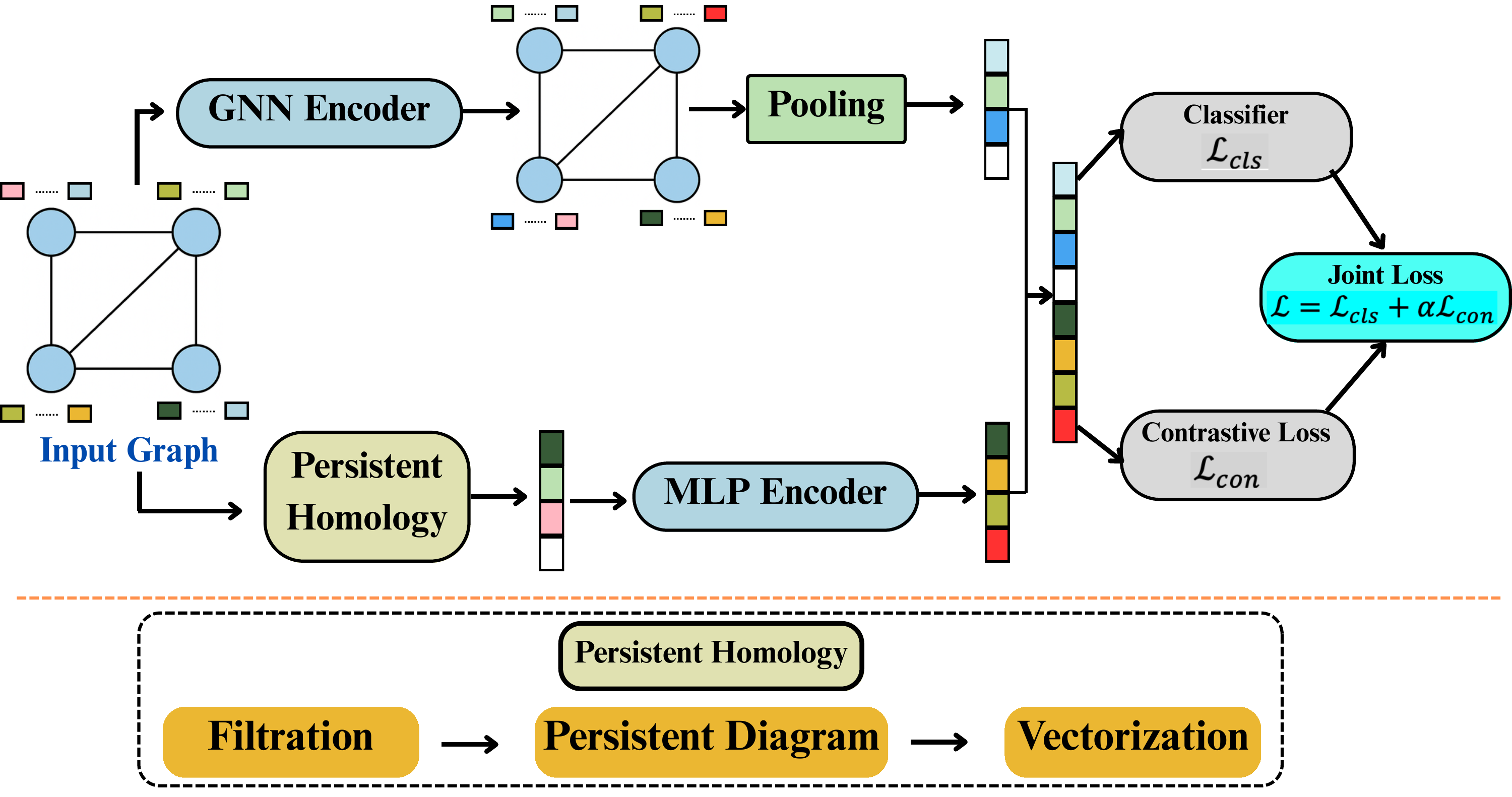}
    \caption{\textbf{Overview of the GraphTCL framework}. Given an input graph, the structural branch encodes node features through a GNN and global pooling, while the topological branch extracts topological features via persistent homology, followed by an MLP encoder. The two embeddings are aligned through a cross–view contrastive loss and jointly optimized with the classification objective.}
    \label{fig:pipeline}
    \vspace{-.2in}
\end{figure}

\paragraph{Joint Training.}  
The overall objective is a weighted sum of the supervised classification loss and the contrastive alignment loss:
\[
\mathcal{L} = \mathcal{L}_{\text{cls}} + \alpha \, \mathcal{L}_{\text{con}},
\]
where $\mathcal{L}_{\text{cls}}$ is the cross-entropy loss on predicted logits and $\alpha$ controls the influence of contrastive learning. In our experiments, we set $\alpha = 0.1$. The overall pipeline of  our model is shown in Figure~\ref{fig:pipeline}.


\section{Experiments}
\subsection{Datasets}
To comprehensively evaluate the effectiveness and generalization capability of our proposed framework, we perform experiments on a diverse collection of benchmark datasets spanning multiple domains. 
We first consider seven widely used datasets from the TU benchmark collection~\cite{morris2020tudataset}, which include \tdata{MUTAG}, \tdata{BZR}, \tdata{PTC}, \tdata{COX2} (molecular graphs) and \tdata{PROTEINS} (Bio-informatics), as well as \tdata{IMDB-BINARY} and \tdata{IMDB-MULTI} (social interaction networks). 
To further examine the scalability and robustness of our model on large and chemically complex structures, we employ five molecular datasets from the Open Graph Benchmark (OGB)~\cite{hu2020open}: \tdata{ogbg-molbace}, \tdata{ogbg-molclintox}, \tdata{ogbg-molbbbp}, \tdata{ogbg-molhiv}, and \tdata{ogbg-molsider}. 
These datasets introduce substantial challenges such as large graph sizes, diverse feature spaces, and label imbalance, thus providing a rigorous testbed for evaluating model generalization in complex molecular domains. The statistics of the data set are shown in Table~\ref{tab:TU} and \ref{tab:OGBG}.

\subsection{Implementation Details}
We evaluate all models under consistent experimental settings to ensure a fair comparison across datasets and domains. For the TU benchmark datasets, we follow the standard practice of using 10-fold cross-validation and report the mean accuracy and standard deviation over the 10 folds. For the OGB molecular benchmarks, we adopt the official scaffold-based data splits provided by the Open Graph Benchmark~\cite{hu2020open}. The model is trained with the Adam optimizer (learning rate $0.001$) on the TU benchmark datasets and the PESG optimizer (learning rate $0.1$) on the OGB benchmarks. The hidden dimensionality of both the GNN and MLP encoders is set to $128$ on the TU benchmarks and $384$ on the OGB molecular benchmarks.

For the graph encoder, we use a Graph Isomorphism Network (GIN)~\cite{xu2019how} or Graph Convolutional Network (GCN)~\cite{kipf2017semi}, with 3 layers on TU datasets and 5 layers on OGB datasets, to aggregate local structural information via message passing. The topological encoder is implemented as a 2-layer MLP that processes the persistent homology (PH) features. Specifically, we compute PH features using the Heat Kernel Signature (HKS)~\cite{sun2009concise} as the filtration function. The HKS values are computed for 10 scales, with thresholds determined by quantile-based sampling to capture representative topological transitions across multiple resolutions. All experiments are conducted in Python on a single NVIDIA A100 GPU with 167 GB of memory. Our implementation is released publicly to facilitate reproducibility and further research.\footnote{\url{https://anonymous.4open.science/r/GraphTCL-3C1F}}

\subsection{Baselines}

To assess the effectiveness of GraphTCL, we compare GraphTCL with a broad spectrum of graph classification methods that cover classical kernels, standard message passing GNNs, optimal transport based models, and recent topology aware approaches. As classical, non learnable baselines we include the Shortest Path kernel (SP-Kernel)~\cite{borgwardt2005shortest} and the Weisfeiler Lehman subtree kernel (WL-Kernel)~\cite{Shervashidze2011WLKernel}. Among supervised GNNs, we consider GCN~\cite{kipf2017semi}, GIN~\cite{xu2019how}, and GAT~\cite{Velickovic2018GAT} as representative message passing architectures, as well as ChebyNet~\cite{defferrard2016convolutional}, APPNP~\cite{gasteiger2019appnp}, GPRGNN~\cite{chien2021adaptive}, and DropGIN~\cite{papp2021dropgnn}, which provide stronger propagation or regularization mechanisms. We further include DIFFPOOL~\cite{ying2018hierarchical} as a hierarchical pooling architecture that learns graph level representations through differentiable clustering. To cover optimal transport based structural matching, we evaluate OT-GNN~\cite{becigneul2020optimal} and TFGW~\cite{vincentcuaz2022template}, which use Wasserstein or fused Gromov Wasserstein distances to compare graphs in a learned latent space. Finally, we compare against two recent methods that are most closely related to our work: TopoGCL~\cite{Chen2024TopoGCL}, a contrastive learning framework that injects topological priors into GNNs via topological augmentations, and PGOT~\cite{qian2024reimagining}, a prototype based classifier built on fused Gromov Wasserstein distances.  We use the same data splits (10-fold CV) and evaluation protocol across all baselines to ensure fair comparison.

\subsection{Results} \label{sec:results}

Table~\ref{tab:TU} reports the mean test accuracy of GraphTCL and competing methods on seven TU graph classification benchmarks. GraphTCL achieves the best performance on six datasets and the second-best on the remaining one, indicating that aligning structural and topological views yields consistent gains across molecular and social graphs. On \tdata{MUTAG}, GraphTCL outperforms the strongest transport-based baselines (PGOT$_1$, OT-GNN) and standard GNNs such as GIN and GAT. The improvements are particularly notable on \tdata{PTC} and \tdata{BZR}, where GraphTCL attains $75.30\%$ and $93.86\%$ accuracy, respectively, surpassing the best GNN baseline (GIN) and the topological contrastive method TopoGCL by up to $6\%$. On \tdata{COX2} and \tdata{PROTEINS}, GraphTCL reaches $89.08\%$ and $80.87\%$ accuracy, respectively, providing clear gains over both message-passing and optimal-transport-based models (TopoGCL, PGOT$_1$). For the social benchmarks \tdata{IMDB-B} and \tdata{IMDB-M}, GraphTCL attains accuracies of $75.80\%$ and $52.40\%$, respectively. On \tdata{IMDB-B}, this represents a modest yet consistent improvement over TopoGCL and all kernel- and GNN-based baselines, whereas on \tdata{IMDB-M}, TopoGCL remains slightly ahead and GraphTCL nonetheless achieves the second-highest accuracy among all competing methods.

\begin{table}[t]
\centering
\setlength{\tabcolsep}{3pt}
\resizebox{\textwidth}{!}{
\begin{tabular}{lccccccc}
\toprule
\textbf{Method} & \textbf{MUTAG} & \textbf{BZR} & \textbf{PTC} & \textbf{COX2} & \textbf{PROTEINS} & \textbf{IMDB-B} & \textbf{IMDB-M} \\
\midrule
\# Graphs &          188 &          405 &          344 &          467 &         1113 & 1000 & 1500 \\
\# Classes &          2 &          2 &          2 &          2 &         2 & 2 &3 \\
\midrule
 SP-Kernel & 69.01\scriptsize{$\pm$5.23} & 77.51\scriptsize{$\pm$1.01} & 55.87\scriptsize{$\pm$6.19} & 78.72\scriptsize{$\pm$1.74} & 68.55\scriptsize{$\pm$3.06} & 38.00\scriptsize{$\pm$5.93} & 28.74\scriptsize{$\pm$3.94} \\
WL-Kernel & 85.26\scriptsize{$\pm$7.37} & 78.29\scriptsize{$\pm$1.71} & 52.29\scriptsize{$\pm$8.48} & 78.51\scriptsize{$\pm$1.15} & 59.82\scriptsize{$\pm$0.00} & \teal{\textbf{66.70\scriptsize{$\pm$3.29}}} & 48.53\scriptsize{$\pm$4.18} \\
\midrule
GCN & 87.89\scriptsize{$\pm$5.29} & {86.83\scriptsize{$\pm$4.25}} & 62.57\scriptsize{$\pm$4.13} & 84.26\scriptsize{$\pm$2.89} & 74.11\scriptsize{$\pm$1.79} & 60.60\scriptsize{$\pm$3.44} & 50.47\scriptsize{$\pm$4.38}\\
 GIN & 93.13\scriptsize{$\pm$6.67} & \magenta{\textbf{87.84\scriptsize{$\pm$3.58}}} & \magenta{\textbf{69.38\scriptsize{$\pm$4.38}}}& 84.75\scriptsize{$\pm$4.18}& 76.92\scriptsize{$\pm$2.73}& 70.40\scriptsize{$\pm$3.32 }& 50.20\scriptsize{$\pm$3.45}\\
GAT & 87.37\scriptsize{$\pm$4.21} & 83.66\scriptsize{$\pm$4.09} & 63.14\scriptsize{$\pm$5.92} &\teal {\textbf{85.11\scriptsize{$\pm$2.13}}} & \teal{\textbf{74.38\scriptsize{$\pm$1.33}}} & 61.70\scriptsize{$\pm$1.49} & 49.93\scriptsize{$\pm$3.09} \\
\midrule
DropGIN & 85.79\scriptsize{$\pm$9.72} & 77.80\scriptsize{$\pm$2.55} & 55.14\scriptsize{$\pm$8.86} & 78.72\scriptsize{$\pm$0.95} & 55.14\scriptsize{$\pm$8.86} & 55.00\scriptsize{$\pm$3.71} & 37.33\scriptsize{$\pm$3.16} \\
DIFFPOOL & 87.50\scriptsize{$\pm$3.35} & 51.83\scriptsize{$\pm$2.17} & 45.75\scriptsize{$\pm$5.25} & 59.67\scriptsize{$\pm$3.32} & 70.60\scriptsize{$\pm$2.74} & 58.71\scriptsize{$\pm$1.48} & 42.08\scriptsize{$\pm$1.93} \\
ChebyNet & 85.79\scriptsize{$\pm$7.08} & 85.61\scriptsize{$\pm$3.69} & 60.57\scriptsize{$\pm$4.20} & 81.06\scriptsize{$\pm$5.58} & 72.68\scriptsize{$\pm$3.25} & 61.80\scriptsize{$\pm$2.32} & 50.20\scriptsize{$\pm$2.21} \\
APPNP & 87.37\scriptsize{$\pm$3.49} & 82.44\scriptsize{$\pm$2.13} & 61.71\scriptsize{$\pm$8.40} & 78.95\scriptsize{$\pm$0.67} & 71.16\scriptsize{$\pm$2.68} & 60.50\scriptsize{$\pm$2.69} & 50.13\scriptsize{$\pm$1.73} \\
GPRGNN & 91.05\scriptsize{$\pm$2.41} & 84.88\scriptsize{$\pm$2.39} & 62.57\scriptsize{$\pm$5.49} & 78.72\scriptsize{$\pm$0.00} & 68.30\scriptsize{$\pm$4.76} & 61.90\scriptsize{$\pm$2.84} & 49.80\scriptsize{$\pm$2.78} \\
OT-GNN & \textcolor{teal}{\textbf{94.74\scriptsize{$\pm$5.77}}} & 85.85\scriptsize{$\pm$3.75} & 62.00\scriptsize{$\pm$9.39} & 84.26\scriptsize{$\pm$2.89} & 72.59\scriptsize{$\pm$3.75} & 61.50\scriptsize{$\pm$3.77} & 49.00\scriptsize{$\pm$1.48} \\
TFGW & 93.68\scriptsize{$\pm$2.11} & 84.39\scriptsize{$\pm$2.93} & 61.14\scriptsize{$\pm$5.74} & 83.19\scriptsize{$\pm$3.08} & 74.29\scriptsize{$\pm$2.14} & 62.20\scriptsize{$\pm$2.18} & \teal{\textbf{50.67\scriptsize{$\pm$2.92}}} \\
TopoGCL & 90.09{\scriptsize $\pm$0.93} & \teal{\textbf{87.17{\scriptsize $\pm$0.83}}} & \teal{\textbf{63.43{\scriptsize $\pm$1.13}}}& 81.45{\scriptsize $\pm$0.55}  & \magenta{\textbf{77.30{\scriptsize $\pm$0.89}}}& \magenta{\textbf{74.67{\scriptsize $\pm$0.32}}} & \blue{\textbf{{52.81{\scriptsize $\pm$0.31}}}} \\
PGOT$_1$ & \textcolor{magenta}{\textbf{95.79\scriptsize{$\pm$3.16}}} & 84.15\scriptsize{$\pm$4.26} & 63.43\scriptsize{$\pm$7.65} & \magenta{\textbf{85.74\scriptsize{$\pm$3.44}}} & 71.52\scriptsize{$\pm$2.27} & 60.60\scriptsize{$\pm$4.94} & 43.07\scriptsize{$\pm$5.54} \\
\midrule
GraphTCL & \textcolor{blue}{\textbf{98.42\scriptsize{$\pm$2.41}}} & \blue{\textbf{93.86\scriptsize{$\pm$3.82}}} & \blue {\textbf{75.30\scriptsize{$\pm$3.64}}}&\blue {\textbf{89.08\scriptsize{$\pm$3.73}}} &\blue{\textbf{80.87\scriptsize{$\pm$3.80}}} &\blue{\textbf{75.80\scriptsize{$\pm$2.68}}} & \magenta{\textbf{52.40\scriptsize{$\pm$2.70}}}\\
\bottomrule
\end{tabular}}
\caption{Test Accuracy ($\%$, mean $\pm$ std) of graph classification of different methods on several widely used benchmarks. The best result is highlighted in \blue{blue}, the second-best is \magenta{magenta}, and the third-best is shown in \teal{teal}.}
\label{tab:TU}
\vspace{-.2in}
\end{table}

\begin{table}[h!]
\centering
\setlength{\tabcolsep}{4pt}
\resizebox{\textwidth}{!}{
\begin{tabular}{lcccccc}
\toprule
{} & \textbf{molsider} & \textbf{molclintox} & \textbf{molbace} & \textbf{molbbbp} & \textbf{molhiv} & \textbf{Avg.Imp} \\
\midrule
\# Graphs &          1427 &          1477 &          1513 &          2039 &         41127 &  {} \\
\# Task &          27 &          2 &          1 &          1 &         1 &  {} \\
\midrule
GCN      &  57.31{\scriptsize$\pm0.95$} &  88.50{\scriptsize$\pm3.79$} &  74.41{\scriptsize$\pm4.21$} &  67.51{\scriptsize$\pm1.67$} &  75.43{\scriptsize$\pm1.91$} & -- \\
GTCL-GCN  &  \textbf{60.67{\scriptsize$\pm2.46$}} &  \textbf{90.52{\scriptsize$\pm1.95$}} &  \textbf{84.85{\scriptsize$\pm0.67$}} &  \textbf{70.31{\scriptsize$\pm1.07$}} &  \textbf{76.14{\scriptsize$\pm1.76$}} &  3.87 \\
\midrule
GIN      &  57.78{\scriptsize$\pm1.36$} &  \textbf{85.15{\scriptsize$\pm3.20$}} &  72.10{\scriptsize$\pm5.18$} &  63.27{\scriptsize$\pm4.05$} &  74.36{\scriptsize$\pm1.89$} & -- \\
GTCL-GIN  &  \textbf{61.76{\scriptsize$\pm1.30$}} &  84.73{\scriptsize$\pm1.13$} &  \textbf{83.22{\scriptsize$\pm3.21$}} &  \textbf{66.73{\scriptsize$\pm2.03$}} &  \textbf{74.83{\scriptsize$\pm1.84$}} &  3.72 \\
\bottomrule 
\end{tabular}}
\caption{\textbf{OGBG datasets.} Performance comparison on OGB molecular graph classification benchmarks. The last column reports the average improvement achieved by our GraphTCL variants over their corresponding GCN and GIN baselines.}
\label{tab:OGBG}
\vspace{-.2in}
\end{table}

We further evaluate our model on five OGBG molecular benchmarks, including the large-scale dataset \tdata{molhov}, as shown in Table~\ref{tab:OGBG}. Using GCN and GIN as backbone baselines, we compare them against the corresponding variants of our method, GraphTCL-GCN and GraphTCL-GIN. Our approach yields consistent improvements across all OGBG datasets, with average gains of $3.87$ and $3.72$ percentage points over GCN and GIN, respectively. These results demonstrate that combining topology and structure via cross-view contrastive learning provides robust benefits across diverse benchmarks and can be readily integrated with standard graph learning architectures.



\begin{table}[t]
\centering
\setlength{\tabcolsep}{3pt}
\resizebox{\textwidth}{!}{
\begin{tabular}{lccccccc}
\toprule
\textbf{Method} & \textbf{MUTAG} & \textbf{BZR} & \textbf{PTC} & \textbf{COX2} & \textbf{PROTEINS} & \textbf{IMDB-B} & \textbf{IMDB-M} \\
\midrule
Topo & 95.76\scriptsize{$\pm$5.16} & 87.67\scriptsize{$\pm$2.37} & 69.47\scriptsize{$\pm$5.07} & 83.30\scriptsize{$\pm$2.49} & 78.17\scriptsize{$\pm$3.39} & 75.40\scriptsize{$\pm$3.14} & 51.47\scriptsize{$\pm$2.51} \\
GIN & 93.13\scriptsize{$\pm$6.67} & 87.84\scriptsize{$\pm$3.58} & 69.38\scriptsize{$\pm$4.38} & 84.75\scriptsize{$\pm$4.18} & 76.92\scriptsize{$\pm$2.73} & 70.40\scriptsize{$\pm$3.32} & 50.20\scriptsize{$\pm$3.45} \\
Topo-GIN & 96.29\scriptsize{$\pm$3.38} & 92.85\scriptsize{$\pm$4.18} & 71.81\scriptsize{$\pm$3.12} & 87.79\scriptsize{$\pm$3.84} & 79.61\scriptsize{$\pm$2.65} & 74.60\scriptsize{$\pm$3.20} & 51.67\scriptsize{$\pm$2.43} \\
\midrule
GraphTCL & \textbf{98.42\scriptsize{$\pm$2.41}} & \textbf{93.86\scriptsize{$\pm$3.82}} & \textbf{75.30\scriptsize{$\pm$3.64}} & \textbf{89.08\scriptsize{$\pm$3.73}} & \textbf{80.87\scriptsize{$\pm$3.80}} & \textbf{75.80\scriptsize{$\pm$2.68}} & \textbf{52.40\scriptsize{$\pm$2.70}} \\
\bottomrule
\end{tabular}}
\caption{\footnotesize \textbf{Ablation on structural and topological components.}
Test accuracy (\%, mean $\pm$ std) of four variants on seven graph benchmarks.
\emph{Topo} and \emph{GIN} use only topological or structural encoders, \emph{Topo-GIN} concatenates both embeddings, and \emph{GraphTCL} further applies cross view contrastive alignment.}
\label{tab:ablation}
\vspace{-.25in}
\end{table}

\subsection{Ablation Studies}

To understand the roles of the structural encoder, the topological encoder, and cross–view contrastive training, we compare four variants in Table~\ref{tab:ablation}: \textbf{Topo}, which uses only the PH-based encoder with an MLP classifier; \textbf{GIN}, which uses only the message-passing GIN encoder; \textbf{Topo-GIN}, which concatenates Topo and GIN embeddings before an MLP; and \textbf{GraphTCL}, the full model with cross–view contrastive alignment between structural and topological embeddings. Topo is already competitive with GIN on several benchmarks, indicating that global connectivity patterns captured by persistent homology are highly discriminative, whereas GIN performs better when local structural motifs dominate. Their complementary behavior suggests that topology and structure encode partially non-overlapping information. The simple \textbf{Topo-GIN} hybrid consistently improves over both single-view baselines, showing that combining structural and topological features is beneficial even without explicit interaction. However, \textbf{GraphTCL} uniformly outperforms Topo-GIN on all datasets, despite using the same encoders and a comparable number of parameters, highlighting the importance of cross–view contrastive alignment.

\begin{wrapfigure}[12]{r}{2in}
\vspace{-.3in}
  \begin{center}
     \includegraphics[width=1.0\linewidth]{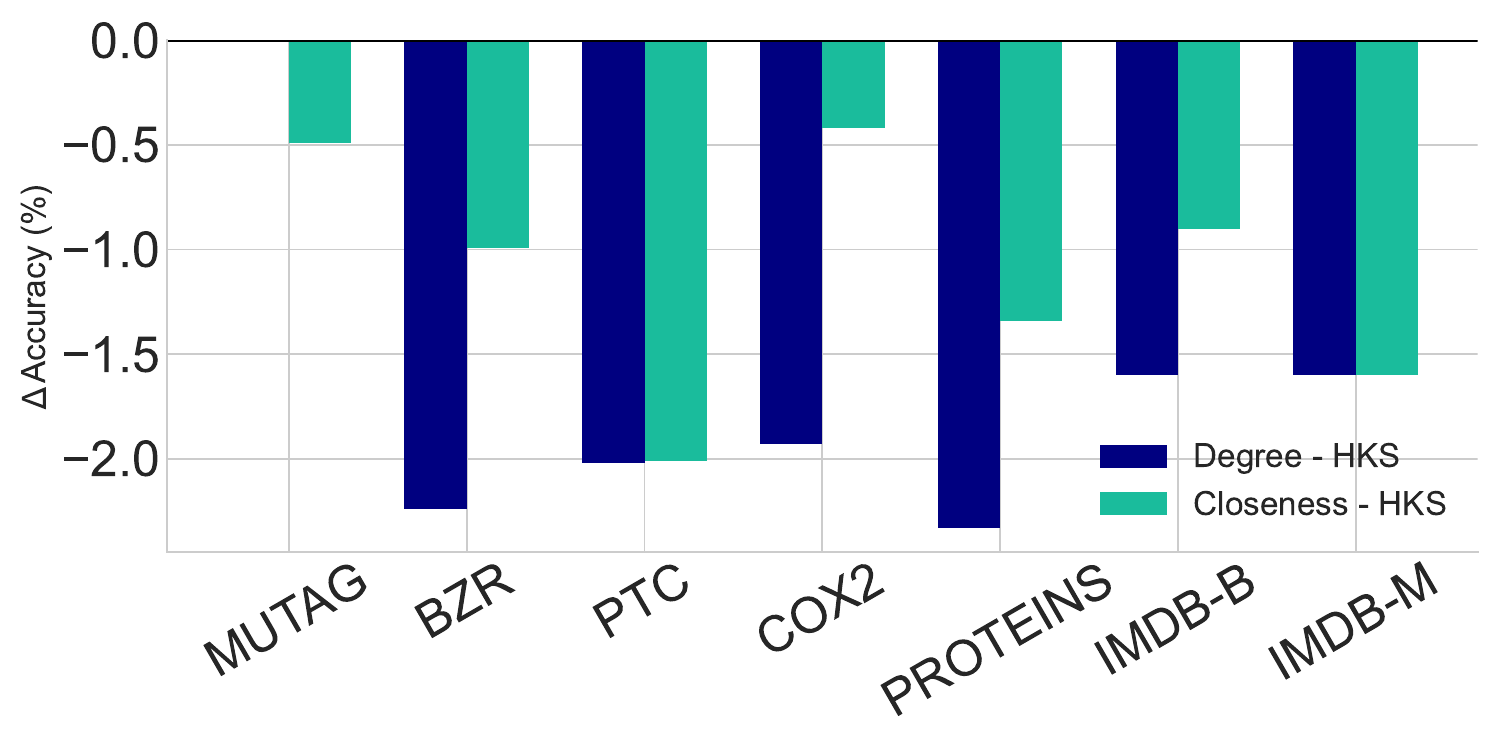}
     \vspace{-.3in}
      \caption{ Relative accuracy drop when replacing the HKS-based filtration with degree- or closeness-based filtrations across seven TU datasets} 
    \label{fig:rel_drop}
      \end{center}
      \vspace{-.2in}
\end{wrapfigure}
Empirically, GraphTCL yields the best accuracy on every benchmark, with the largest gains appearing on datasets where the two modalities are most complementary (e.g., \tdata{MUTAG}, \tdata{BZR}, and \tdata{PTC}). These results confirm that (i) topological features are individually strong, (ii) naive fusion already helps, but (iii) explicitly \emph{regularizing} the structural representation with a topological view via cross–view contrastive learning is crucial to fully exploit their synergy.

In our architecture, we use HKS as the default filtration to extract topological signatures, but GraphTCL is compatible with any structural filtration. To verify this, we replace HKS with degree and closeness centrality as alternative filtrations. As shown in Figure~\ref{fig:rel_drop}, the relative performance drop is at most $2\%$ across all datasets, demonstrating that GraphTCL is stable with respect to the choice of filtration. At the same time, domain-specific filtrations (such as HKS in molecular graphs) remain important for maximizing performance.

\smallskip

\noindent \textbf{Key takeaways.} \quad Taken together, the TU and OGB experiments support three main conclusions. First, persistent-homology-based features alone are already competitive with strong message-passing GNNs, which confirms that global connectivity patterns carry substantial predictive signal. Second, simple feature-level fusion of structural and topological embeddings improves over either modality in isolation, indicating that they encode complementary information. Third, GraphTCL, which adds a cross-view contrastive objective on top of the same encoders, yields the most consistent gains across datasets while keeping the architecture simple and backbone agnostic. This suggests that explicitly encouraging agreement between structural and topological representations is a principled and practical way to inject topology into graph learning, rather than treating it only as an auxiliary regularizer or handcrafted feature. Our current study is limited to graph-level classification with precomputed PH features and fixed filtrations, and extending GraphTCL to learned filtrations and node- or edge-level tasks is a promising direction for future work.

\section{Conclusion}

We introduced GraphTCL, a supervised cross view contrastive framework that jointly learns structural and topological representations for graph classification by aligning a GNN encoder with a persistent homology based encoder in a shared latent space. This design lets the structural view benefit from global connectivity cues while guiding the topological view toward task relevant patterns. Across seven benchmarks, GraphTCL consistently matches or surpasses strong kernel, GNN, transport, and topological baselines, and our ablation study shows that while topological features and simple hybrids (Topo-GIN) are already competitive, explicit cross view alignment is key to fully exploiting their synergy. Future work includes extending GraphTCL to richer topological summaries and dynamic or heterogeneous graphs, and developing a theory that explains when and why cross view alignment improves generalization in graph learning.


%
%
%

\bibliographystyle{splncs04}
\bibliography{Ref}

\end{document}